\documentclass[conference]{IEEEtran}
\usepackage{amsmath}
\usepackage{amsmath,amssymb,amsfonts}
\usepackage{colortbl}
\usepackage[table,xcdraw]{xcolor}
\usepackage{graphicx}
\usepackage{cite}
\ifCLASSINFOpdf
\else
\fi
\begin{document}

%
\title{3D ResNet with Ranking Loss Function for Abnormal Activity Detection in Videos}

\author{\IEEEauthorblockN{Shikha Dubey}
\IEEEauthorblockA{School of Electrical Engineering\\and Computer Science\\
Gwangju Institute of Science and \\ Technology\\
Gwangju 61005, Republic of Korea\\
Email: shikhad.bhu@gmail.com}
\and
\IEEEauthorblockN{Abhijeet Boragule}
\IEEEauthorblockA{School of Electrical Engineering\\and Computer Science\\
Gwangju Institute of Science and \\ Technology\\
Gwangju 61005, Republic of Korea\\
Email: abhijeet@gist.ac.kr}
\and
\IEEEauthorblockN{Moongu Jeon}
\IEEEauthorblockA{School of Electrical Engineering\\and Computer Science\\
Gwangju Institute of Science and \\ Technology\\
Gwangju 61005, Republic of Korea\\
Email: mgjeon@gist.ac.kr}}


%


\maketitle

\begin{abstract}
Abnormal activity detection is one of the most challenging tasks in the field of computer vision. This study is motivated by the recent state-of-art work of abnormal activity detection, which utilizes both abnormal and normal videos in learning abnormalities with the help of multiple instance learning by providing the data with video-level information. In the absence of temporal-annotations, such a model is prone to give a false alarm while detecting the abnormalities. For this reason, in this paper, we focus on the task of minimizing the false alarm rate while performing an abnormal activity detection task. The mitigation of these false alarms and recent advancement of 3D deep neural network in video action recognition task collectively give us motivation to exploit the 3D ResNet in our proposed method, which helps to extract spatial-temporal features from the videos. Afterwards, using these features and deep multiple instance learning along with the proposed ranking loss, our model learns to predict the abnormality score at the video segment level. Therefore, our proposed method 3D deep Multiple Instance Learning with ResNet (MILR) along with the new proposed ranking loss function achieves the best performance on the UCF-Crime benchmark dataset, as compared to other state-of-art methods. The effectiveness of our proposed method is demonstrated on the UCF-Crime dataset.    
\end{abstract}


%
\IEEEpeerreviewmaketitle

\section{Introduction}
To accomplish human security in a more vigorous way and to obviate many crimes in society, many surveillance cameras have been installed in numerous places such as, shopping complexes, highways, roads, banks, etc. Due to the increasing number of surveillance cameras, plenty of human operators are needed to monitor for any kind of abnormal activities in these videos. Moreover, the demands are laborious and time consuming. Since, for a human being, it is tedious to look at the surveillance videos all the time and additionally, watching multiple videos simultaneously can lead to miss detect the abnormal activity. Therefore, to overcome from all these troubles and to make human work more efficient, an automated abnormal activity detection system is essential, which can generate some signal for any abnormal activity happening in the videos automatically. Therefore, accomplishing abnormal activity detection task with a low false alarm rate and with high accuracy is a challenging and critical task in the computer vision field.

Several algorithms \cite{A2,A3,A6,A9,A11,A15,A16,A17,A18,A19,A20,A21} etc. have been proposed to detect abnormal activities in videos. Recently, W. Sultani \textit{et al.} \cite{A6}, has proposed a new approach to solve all the shortcomings of the traditional approaches. In their method, to detect the abnormal activity, they have trained their model with the help of multiple instance learning (MIL) \cite{A12, A24} by utilizing the weakly labeled training videos. Our method is motivated by their work. In our study, our main goal is to mitigate the false alarm rate in the abnormal activity detection task. In order to achieve our goal and to extract robust spatial-temporal features from the videos, we have proposed a new 3D deep Multiple Instance Learning with ResNet (MILR) neural network model, which consists of the 3D ResNet \cite{A5,A13} and the deep MIL \cite{A12,A6} method, along with the new proposed ranking loss. Where recent advancement of 3D deep neural network in the video action recognition task motivates us to use 3D ResNet \cite{A5,A13} in the abnormal activity detection task. Extracted features from 3D ResNet and video-level labeling are used to train the model in a weakly-supervised manner with the help of deep MIL \cite{A12, A6} along with the new proposed ranking loss. 

The rest of this paper is organized as follows: Section \ref{RelW}, gives a brief overview of the existing algorithms. Section \ref{pa}, gives the details of our proposed abnormal activity detection method. Detailed implementation and experimentation along with the evaluation of our method are given in Section \ref{IE}. Lastly, Section \ref{con} concludes the paper.

\section{Related Work} \label{RelW}

A brief review of some recent algorithms is presented in this section. Algorithms such as traffic monitoring systems \cite{A2,A3,A20}, abandoned object detector \cite{A7} and violence detection in crowds \cite{A19}, etc., are proposed to detect abnormalities in certain specific tasks. Therefore, generalization of these algorithms for detection of other abnormal activities is difficult. Moreover, listing all possible normal and abnormal activities in real-world scenarios is difficult. Therefore, the abnormal activity detector should be less dependent on the prior knowledge of the activity. In order to fulfill such requirements, sparse coding based approaches \cite{A18,A9} have been proposed, where the extracted features from the normal videos are used to construct the dictionary to represent those normal videos. Moreover, the network is trained only for the normal videos. At the testing time, the high reconstruction error while generating the dictionary indicates the presence of the abnormal activity. Since it is difficult to list all kinds of normal videos, these methods are prone to generate false alarms.

The success of the deep learning methods in the image processing tasks \cite{A5} and action recognition \cite{A14, A13, A5} task, motivated the researchers to apply these methods in the case of the abnormal activity recognition. Consequently, a deep auto-encoder based approach \cite{A11,A15,A16} has been proposed to learn the features of the normal activities automatically, but generalization of these methods for real-world scenarios is difficult.

Moreover, recently a new state-of-art method has been introduced \cite{A6}, in which the detection of the abnormal activity is done with the help of C3D \cite{A1} and MIL \cite{A12,A24}. This method has trained the model with the help of both normal and abnormal videos. Since MIL is basically semi-supervised learning, in which we provide the dataset with the video-level information, such a model is prone to indicate false alarms while detecting the abnormalities. Therefore, our proposed method is motivated by their method. In our method, the features have been extracted from the 3D ResNet \cite{A5,A13} and the network is trained with the help of a new ranking loss function, which reduces the false alarm rate in an abnormal action detection task.

\begin{figure*}

	\begin{center}
		\includegraphics[width=1.0\linewidth]{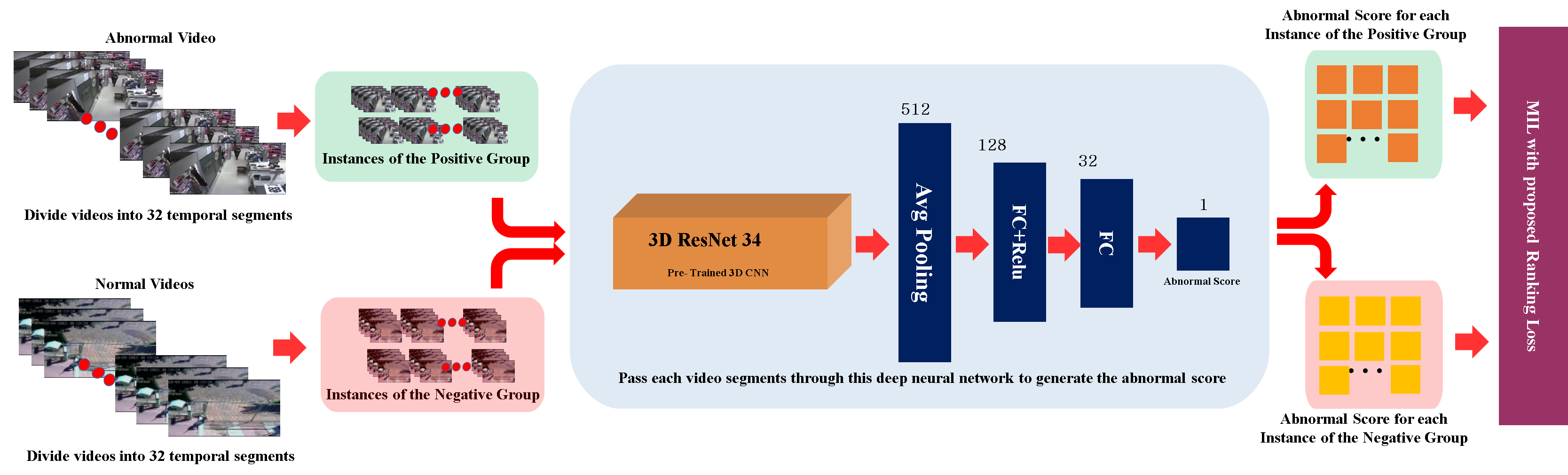}
		
	\end{center}
	\caption{The model of the proposed abnormal activity detection algorithm. Each normal and abnormal videos is divided into groups of positive instances and negative instances. Then the proposed neural network predicts the abnormality score for each instance with the help of MIL along with the new ranking loss.}
	\label{model_figure}
	
\end{figure*}
 
\section{The Proposed Abnormal Activity Detection Algorithm} \label{pa}
The proposed abnormal activity detection algorithm is shown in the Fig. \ref{model_figure}. It is a combination of mainly three steps. First, each video from the training set are divided into groups of positive instances and negative instances by utilizing their video-level labeling. Then, the 3D ResNet-34 \cite{A5,A13}, which is trained on the Kinetics dataset \cite{A23} for the action recognition task, extracts spatial-temporal features from those instances. Next, the extracted features are passed through the deep neural network to predict the abnormality score for each instance of the video. Finally, the network is trained in a weakly-supervised way with the help of the deep MIL \cite{A12,A24} along with the proposed new ranking loss function. All the steps are explained in the following subsections.

\subsection{Feature Extraction through the 3D ResNet} \label{ResNet}
Before extracting features from videos for further processing, first each video from the training set are divided into the equal number of non-overlapping video segments AND Further each segments are grouped into the groups of positive instances and negative instances. In the proposed algorithm, we have taken advantage of the video-level knowledge. The segments from abnormal videos have been grouped into positive instances, and the segments of the normal videos have been grouped into the negative instances. Where, $ C_a $ and $ C_n $ are the representation for positive group and negative group respectively \cite{A6}.

Subsequently, to extract the spatial-temporal features, each instance are passed through the pre-trained 3D ResNet-34 \cite{A5,A13} network, which is pre-trained on the Kinetics dataset \cite{A23}. From the study of S. Tiago \textit{et al.} \cite{A21}, pre-trained CNNs are good feature extractor in the abnormal activity detection task. However, in our algorithm, we have used the 3D CNN, which is a better spatial-temporal extractor in the video processing task. Moreover, from the study of K. Hara \textit{et al.} \cite{A5,A13}, we are confident using 3D ResNet, since it outperforms any other 3D CNNs.
From Figures \ref{ROC_Curve} and \ref{AUC_Graph}, using 3D ResNet-34 in the abnormal activity detection task results in better performance than any other algorithm.

\subsection{Deep Multiple Instance Learning and Proposed Ranking Loss Function}
In this paper, the network is trained in a weakly-supervised manner with the help of the deep multiple instance learning (MIL) \cite{A12,A24}. In the absence of frame-level or temporal-level labels, we have used deep MIL as similar to the work of W. Sultani \textit{et al.} \cite{A6}. By providing video-level labels, MIL can train the network with the help of ranking loss function as  follows.

As discussed in the Section \ref{ResNet}, we have grouped of all instances. Where the group of positive instances is represented as $ C_a $, since we have at least one instance with the abnormal activity in it, and the group of negative instances is represented as $ C_n $. The number of total instances in both groups is equal to $ n $. Since the accurate labels for the instances of the positive group are unknown, we can use the following optimization function \cite{A24} for the binary classification task as below \cite{A24}:

\begin{equation} \label{eq1}
\small {
    \min_{\textbf{w}} \frac{1}{m}\sum_{i=1}^m \overbrace{\max \left(0, 1- Y_{C_i} \left(\max_{k \in C_i} \left(w . \phi \left(x_k\right)\right)-b\right)\right)}^{\widehat{a}} + \frac{1}{2} \parallel \textbf{w} \parallel^2 }
\end{equation}
where $\widehat{a}$ represents the hinge loss function, $Y_{C_i}$ represents the group-label,  $\phi \left(x\right)$ is a feature representation of the instance, $b$ represents the bias term, $ m $ is the total number of the training examples and $\textbf{w}$ represents model weights.

In this study, the abnormal detection problem is treated as a regression problem \cite{A6}, since we would like to assign a higher abnormality score to all the abnormal instances than the normal instances. Therefore, ranking loss could be one of the solutions to train our model. If we have instance-level labeling, the ranking objective function is mentioned as below:
\begin{equation} \label{eq2}
   S\left(I_a\right)>S\left(I_n\right)
\end{equation}
where, $I_a$ and $I_n$ represent the instances of the abnormal video and normal video respectively, functions $ S\left(I_a\right) $ and $ S\left(I_n\right) $ give the predicted abnormal scores for the corresponding instances of the abnormal and the normal video, respectively. The score function $ S(x) $ ranges in between $ 0 $ and $ 1 $.
 
 \textbf{ Proposed Ranking Loss Function:} In the absence of the instance-level labeling, we cannot use the above equation \ref{eq2}. Therefore, we propose a new ranking loss function for training our model using MIL. Before proposing the new loss function, we need to consider the possible false alarm cases in our task. 
 \subsubsection{False Alarm Cases} \label{FalseAlarm}There can be two possible cases of false alarms from our model. 
 
 The \textbf{first case (case 1)} is when our model predicts normal activity as an abnormal activity, which is the case of false positive. The \textbf{second case (case 2)} is when our model predicts abnormal activity as a normal activity, which is the case of false negative. 
 \subsubsection{Proposed Loss Function} \label{ProposedLoss}As our task is to mitigate all kinds of false alarm rates as described in \ref{FalseAlarm}, in order to reach our goal, following ranking conditions has been proposed:  
 
 \begin{equation} \label{eq3}
    \max_{i\in C_a} S\left(I_a^i\right)  >  \max_{ i\in C_n} S\left(I_n^i\right) 
\end{equation}

\begin{equation} \label{eq4}
    \max_{ i\in C_a} S\left(I_a^i\right)  >  \min_{ i\in C_a} S\left(I_a^i\right)
\end{equation}

Equations \ref{eq3} and \ref{eq4} are to avoid case 1 of the false alarm, when the model predicts a normal instance as an abnormal instance. Equation \ref{eq3} compares the maximum ranked instances from each group \cite{A6}, where the maximum ranked instance from the positive group is most likely to be the true positive and the maximum ranked instance from the negative group can be the case of false positive. 

Equation \ref{eq4} compares the maximum ranked instance and minimum ranked instance from the positive group, where the maximum ranked instance from the positive group is most likely to be the true positive and the minimum ranked instance from the positive group can be the case of false positive.

As mentioned below, arrange the instances of positive group in the descending order of the abnormality score:
\begin{equation} \label{eq5}
    [M_1,M_2,M_3, ... , M_{n}]=order_{ i\in C_a}^{desc} S\left(I_a^i\right)
\end{equation}
where $n$ is the total number of the instances in each group.

\begin{equation} \label{eq6}
    M_2\left(S\left(I_a\right)\right) >  \max_{ i\in C_n} S\left(I_n^i\right) 
\end{equation}
\begin{equation}  \label{eq7}
    M_3 \left(S\left(I_a\right)\right)  >  \max_{ i\in C_n} S\left(I_n^i\right) 
\end{equation}
Equations \ref{eq6} and \ref{eq7} are used to avoid both cases 1 and 2 of the false alarms. Equations \ref{eq6} and \ref{eq7} compare the second and third ranked instances of the positive group with the maximum ranked instances of the negative group, respectively. 

We have proposed these ranking equations, since the videos of the training dataset are large in size, and there can be multiple instances of the abnormal activity in the video. Therefore, we need to maximize abnormality scores of the other instances of the positive group to avoid the case $2$ of the false alarm; where the model predicts the abnormal instance as a normal instance. Moreover, we need to minimize the abnormality score of all the instances of the negative group.

We don't require any instance-level labeling from equation \ref{eq3} to equation \ref{eq7}. Therefore, in terms of the predicted abnormality scores we would like to make all the negative instances far apart from all the positive instances. Moreover, in order to satisfy all the ranking conditions from equation \ref{eq3} to equation \ref{eq7}, the new proposed ranking loss function is given as below in the hinge loss formulation:

\begin{multline}\label{eq8}
    l\left(C_a,C_n\right)= \\ l_1\left(C_a,C_n\right)+l_2\left(C_a,C_a\right)+l_3\left(C_a,C_n\right)+ l_4\left(C_a,C_n\right) 
\end{multline}
where, $l_1\left(C_a,C_n\right), l_2\left(C_a,C_a\right), l_3\left(C_a,C_n\right)$ and $l_4\left(C_a,C_n\right) $ are defined as below:
\begin{equation}\label{eq9}
   l_1\left(C_a,C_n\right) = \max \left(0,1-\max_{ i\in C_a} S\left(I_a^i\right)+\max_{ i\in C_n} S\left(I_n^i\right)\right)
\end{equation}
\begin{equation}\label{eq10}
  l_2\left(C_a,C_a\right) = \max \left(0,1-\max_{ i\in C_a} S\left(I_a^i\right)+\min_{ i\in C_a} S\left(I_a^i\right)\right)
\end{equation}
\begin{equation}\label{eq11}
   l_3\left(C_a,C_n\right) = \max \left(0,1-M_2\left(S\left(I_a\right)\right)+\max_{ i\in C_n} S\left(I_n^i\right)\right)
\end{equation}
\begin{equation}\label{eq12}
    l_4\left(C_a,C_n\right) = \max \left(0,1-M_3\left(S\left(I_a\right)\right)+\max_{ i\in C_n} S\left(I_n^i\right)\right)
\end{equation}

As in the study of W. Sultani \textit{et al.} \cite{A6}, to maintain the temporal smoothness of the abnormality score and to maintain the sparsity of the abnormal scores, two constraints have been proposed as below:
\begin{equation} \label{eq13}
    temporal_{constraint}= \mu_1 \sum_i^{n-1} \left(S\left(I_a^i\right)-S\left(I_a^{i+1}\right)\right)^2
\end{equation}
\begin{equation} \label{eq14}
    sparsity_{constraint}=\mu_2 \sum_i^{n} S\left(I_n^i\right)
\end{equation}

Now, the proposed ranking loss can be written as: 
\begin{multline}\label{eq15}
    l\left(C_a,C_n\right)= \\ l_1\left(C_a,C_n\right)+l_2\left(C_a,C_a\right)+l_3\left(C_a,C_n\right)+ l_4\left(C_a,C_n\right) +\\ 
    temporal_{constraint}+  sparsity_{constraint} 
\end{multline}

Therefore, the final loss function to train our proposed model is given as below:
 
\begin{equation} \label{eq16}
    L\left( \textbf{W} \right) = l\left(C_a,C_n\right) + \mu_3 \parallel \textbf{W} \parallel^2
\end{equation}
where, \textbf{W} are the model weights and $\mu_1, \mu_2$ and $\mu_3$ are the hyper-parameters of the model.

\section{Implementation and Experimentation} \label{IE}

\subsection{Implementation} \label{Impl}
\subsubsection{Dataset} \label{dataset}
There are several standard datasets \cite{A7, A10, A8, A9, A6} are available for the abnormal activity detection task. In UMN dataset \cite{A7}, only one class of abnormality is provided to detect. Subway Exit and Entrance datasets \cite{A10} have only two classes of abnormality to detect. UCSD Ped1 and Ped2 datasets \cite{A8} and Avenue dataset \cite{A9}, consist of some simple abnormal activities, which are less realistic and cannot be generalized for the real-world scenarios. Therefore, in this study, we have used the recently introduced UCF-Crime dataset \cite{A6}. This is the largest and challenging dataset in the abnormal activity detection task. Since this dataset consist of total 1900 long untrimmed real-world videos with 950 normal videos and 950 abnormal videos. Moreover, it has 13 classes of real-world abnormal activities.  

\textbf{Training set and Testing set:} The whole dataset is divided into two sets, training set and testing set \cite{A6}. Both the sets consist videos of all 13 classes of abnormal activities. Training set consists of 810 abnormal videos and 800 normal videos. Where, testing set consists of 140 abnormal videos and 150 normal videos. 
\subsubsection{Details of the Implementation} \label{detImple}
In our implementation, before the feature extraction task, each training and testing videos have been divided into $32$ non-overlapping video segments. Then each video frame is resized into $112*112$ pixel size and their frame rate per second (fps) is fixed to $30$. Then in the \textbf{training phase},  these segments are divided into the group of positive instances and the group of negative instances. Then, 3D ResNet-$34$ \cite{A5, A13} extracts spatial-temporal features for every $16$ frames of the video.As shown in the Fig.~\ref{model_figure}, these extracted $512$D features for each instance are passed through fully connected (FC) neural network of $3$-layers. Where, the first FC layer consists of $128$ units with the ReLU activation function and second FC layer consists of $32$ units and the last FC layer has $1$ unit with the sigmoid activation function. Similar to the W. Sultani \textit{et al.} \cite{A6} study, $60\%$ dropout has been used in between the $3$-FC layers. Our proposed network is trained using Adagrad optimizer with a $0.001$ initial learning rate. For the best performance of our model, the values of all hyper-parameters are set as \cite{A6}, $ \mu_1=\mu_2=8*10^{-5}$ and $\mu_3=0.01$.
Moreover, the batch of $60$ randomly chosen videos ($30$ abnormal and $30$ normal) has been used while training the model. Then the network has been trained for $25,000$ epochs with the help of the deep MIL along with the new proposed ranking loss as given in equation \ref{eq16}.

\subsection{Experimentation} \label{Exper}
\subsubsection {Testing Phase} \label{Test}
In the testing phase, each testing video is divided into $32$ non-overlapping video segments and each video frame is resized into $112*112$ pixel size and their frame rate per second (fps) is fixed to $30$. Then all the segments are passed through our proposed deep neural network, which predicts the abnormality score for each video segment of the video.  
 
\subsubsection{Evaluation Methods} \label{EvalMet}
Similar to the previous abnormal activity detection algorithms \cite{A6, A9, A11} etc., to evaluate the effectiveness of our proposed method, we have also used two quantitative evaluation methods as receiver operating characteristic curve (ROC-Curve) and area under the curve (AUC).

\subsubsection{Comparison with the other State-of-Art Methods}

In this paper, to check the effectiveness of our proposed method, we have compared it with the other three state-of-art methods. The first one is C. Lu \textit{et al.} \cite{A9} algorithm, which uses dictionary learning based approach. The second one is M. Hasan \textit{et al.} \cite{A11} algorithm, which uses deep auto-encoder based approach and the third one is the recently introduced approach by W. Sultani \textit{et al.} \cite{A6}, which uses C3D \cite{A1} network to extract features and to predict the abnormality. Our method is motivated by W. Sultani \textit{et al.} \cite{A6} algorithm. Additionally, SVM binary classifier is used as a baseline algorithm in the comparison.

\begin{figure}
	\begin{center}
    \includegraphics[width=1.0\linewidth]{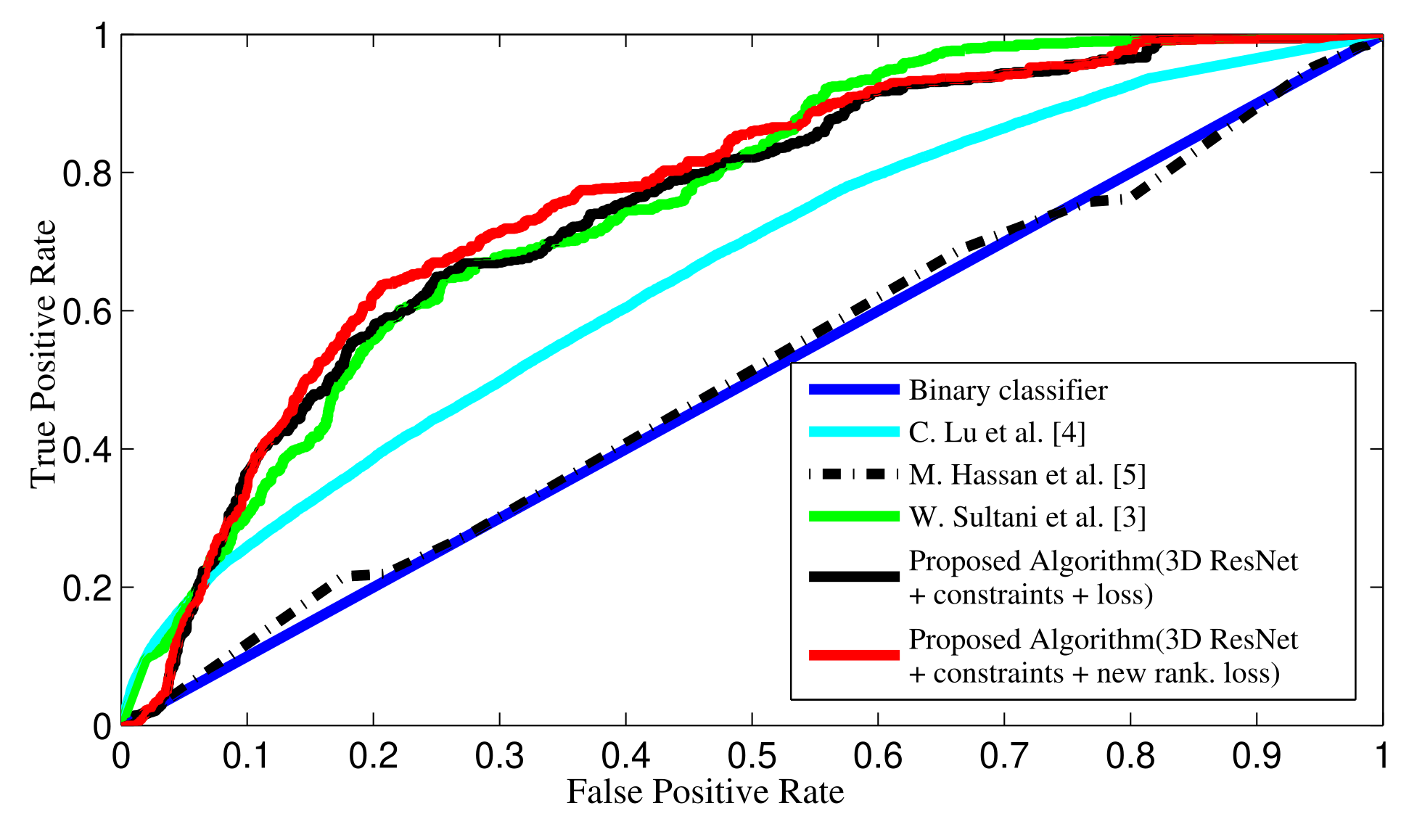}
	\end{center}
	\caption{Comparison of ROC-Curves: blue, dotted-black, cyan, green, black, and red color curves represent binary classifier, M.Hasan \textit{et al.} \cite{A11}, C. Lu \textit{et al.} \cite{A9}, W. Sultani \textit{et al.} \cite{A6}, the proposed method (3D ResNet+constr.+loss) and the proposed method with new ranking loss (3D ResNet+constr.+new rank. loss), respectively.}
	\label{ROC_Curve}
\end{figure}

\begin{figure}

	\begin{center}
		\includegraphics[height= 5.0 cm,width = 8.5 cm]{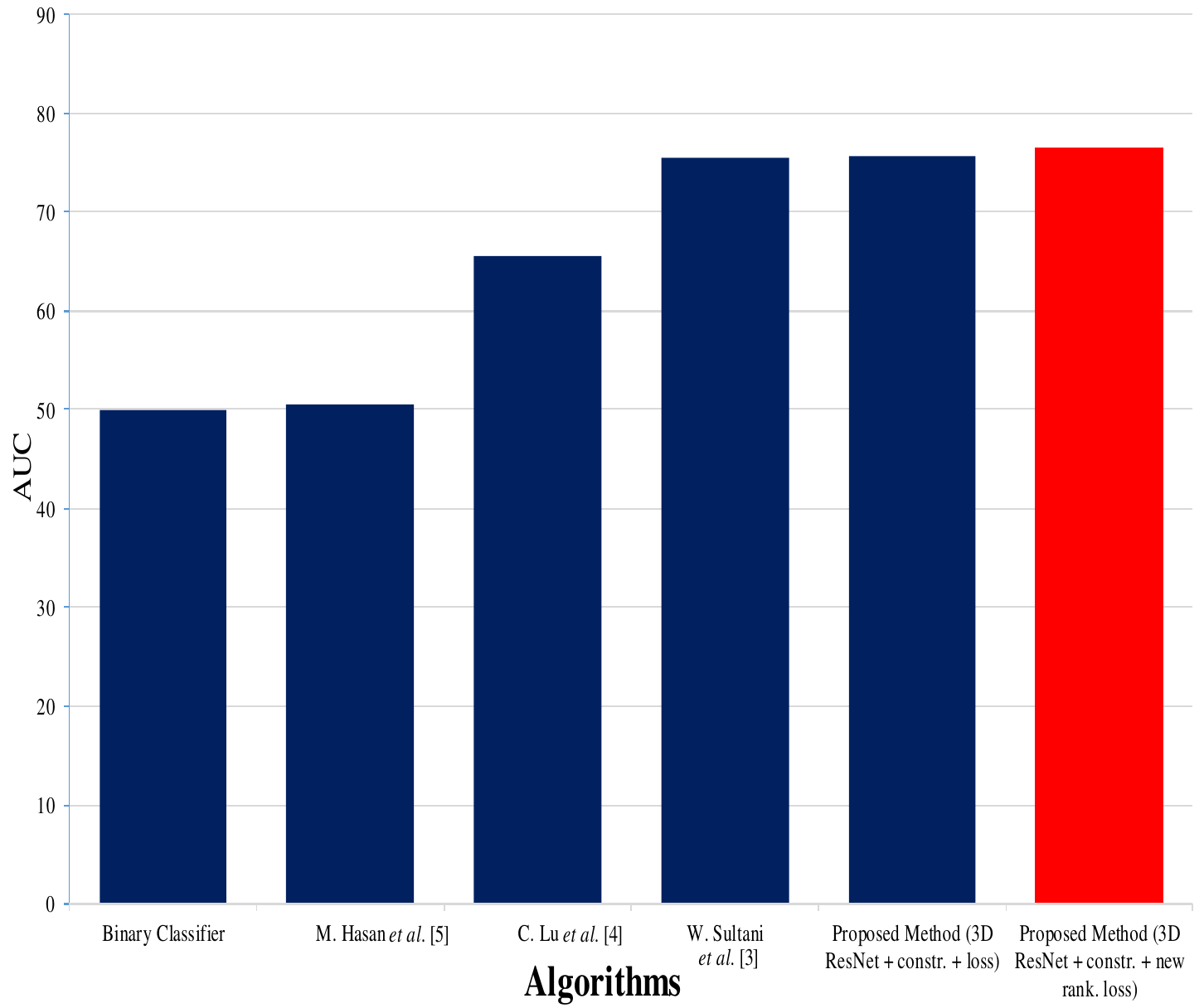}
		
	\end{center}
	\caption{AUC Comparison of all the methods on the UCF-Crime Dataset, X- axis describes the algorithm names and Y-axis describes AUC for each algorithm. Our proposed method outperforms in all.}
	\label{AUC_Graph}
	
\end{figure}

As shown in Figures \ref{ROC_Curve}, \ref{AUC_Graph} and in Table \ref{AUC_table}, ROC-Curve and AUC methods show the \textbf{quantitative comparisons} of our algorithm with other state-of-art methods. Fig.~\ref{ROC_Curve} shows the comparison of all the ROC-Curves and it shows that our proposed method outperforms in all other methods. Similarly from the Table \ref{AUC_table} and Fig.~\ref{AUC_Graph}, we can say that our proposed method gives the highest AUC score in all. Additionally, the effectiveness of our proposed ranking loss function is also shown. From these comparisons, it is shown that our proposed algorithm along with the proposed ranking loss shows the best performance in other state-of-art methods.

\begin{table}
\caption{Comparison of all the methods on the UCF-Crime Dataset.}
\centering
\begin{tabular}{|c|c|}
\hline
\textbf{Algorithms}                     & \textbf{AUC}                         \\ \hline
Binary Classifier                       & 50.0                                 \\ 
M. Hasan \textit{et al.} \cite{A11}     & 50.6                                 \\ 
C. Lu \textit{et al.} \cite{A9}         & 65.51                                \\ 
W. Sultani \textit{et al.} \cite{A6}    & 75.41                                \\ 
Proposed Method (3D ResNet + constr. + loss) &  \textbf{75.62}                        \\ 
Proposed Method (3D ResNet + constr. + new rank. loss)    & {\textbf{76.67}} \\ \hline
\end{tabular}

\label{AUC_table}
\end{table}

\begin{figure*}

	\begin{center}
		\includegraphics[width=1.0\linewidth]{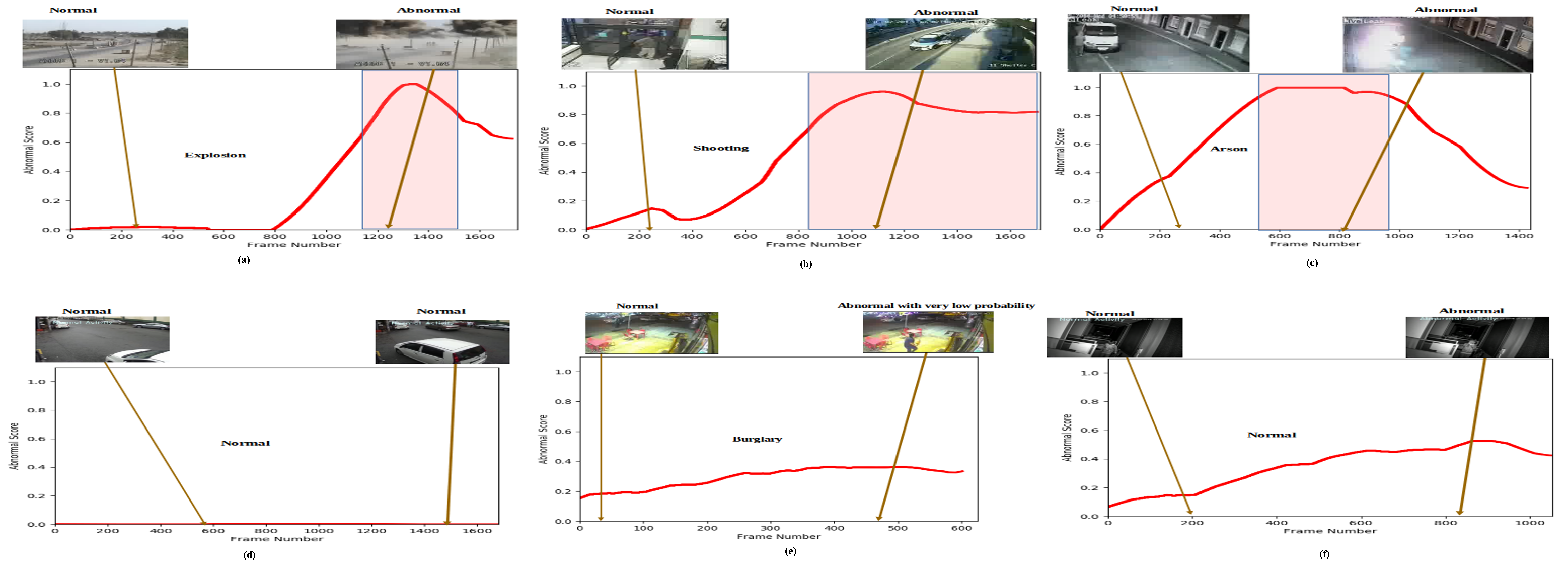}
		
	\end{center}
	\caption{Qualitative Analysis of our proposed algorithm on the testing videos of UCF-Crime dataset. The colored windows show the temporal-region of abnormal activities from ground truth. Where (a), (b), (c) and (e) are results of the abnormal activities named as Explosion, Shooting, Arson and Burglary respectively. (d), (f) are results for the normal videos. Where, (e) and (f) show failure cases of our proposed method.}
	\label{Qualitative_graphs}
\end{figure*}

In Fig.~\ref{Qualitative_graphs}, the \textbf{qualitative analysis} of our proposed algorithm is shown for 6 testing videos, where (a)-(c) show the detection of the abnormal activities correctly with the high abnormality scores for all abnormal video segments and low abnormality scores for all normal video segments, (d) shows the detection of the normal activity accurately with the lowest abnormality scores almost $0$ for all normal video segments. Additionally, (e) and (f) show failure case of our proposed method. This shows our methods get fail in detecting some abnormal activities correctly. From (e) and (f), we can say that it gets fail when there is a lot of light variation.  

\subsubsection{Analysis of the Proposed Algorithm}
\textbf{False alarm rate in normal videos: } Table \ref{false_alarm_nomral_videos} shows the comparison of false alarm rate at the threshold of $ 50\% $ of different algorithms. These false alarm rates have been calculated on the normal videos of the UCF-Crime \cite{A5} testing dataset. Since, normal activities take place more often in real-world scenarios, therefore the detector should give less false alarm rate for normal videos. As shown in Table \ref{false_alarm_nomral_videos}, our proposed method gives the lowest false alarm rate on normal videos as compare to other algorithms. Additionally, the effectiveness of using new proposed ranking loss function is also shown in the Table \ref{false_alarm_nomral_videos}. This comparison shows that using both abnormal and normal videos in training the network improves the abnormal activity detection task and we can generalize our method for other real-world related abnormal activity detection.
\textbf{False alarm rate in abnormal videos: } Since, our method gives the highest AUC in all, therefore in Table \ref{false_alarm_abnormal_videos}, we have just shown the comparison of false alarm rate at the threshold of $ 50\% $ on the basis of variation of our proposed method. Abnormal activities happen rarely in the real-world scenarios, but still detecting the abnormal activity accurately is an important task. Therefore the detector should give less false alarm rate for abnormal videos. As shown in Table \ref{false_alarm_abnormal_videos}, our proposed method along with the new ranking loss function gives the lowest false alarm rate on abnormal videos. The table shows the effectiveness of using the new loss function. 

Therefore, both the tables \ref{false_alarm_nomral_videos} and \ref{false_alarm_abnormal_videos}, indicate that we have reduced both cases of false alarm with the help of our proposed method.

\begin{table}[]
\caption{Comparison of false alarm rate on normal videos of UCF-Crime testing dataset.}
\centering
    \begin{tabular}{|c|c|}
    \hline
   \textbf{Algorithms}                     & \textbf{\begin{tabular}[c]{@{}c@{}}False \\  Alarm Rate\\ (Normal Videos)\end{tabular}} \\ \hline
   
    M. Hasan \textit{et al.} \cite{A11}                    & 27.2    \\
   
   C. Lu \textit{et al.} \cite{A9}                      & 3.1  \\                                                                                    
  W. Sultani \textit{et al.} \cite{A6}                   & 1.9     \\                                                                                  
  Proposed Method(3D ResNet+constr.+loss) & 0.83   \\                                                                                     
   Proposed Method(3D ResNet+constr.+new rank.loss)     & \textbf{0.80}   \\ \hline                                                    
    \end{tabular}

    \label{false_alarm_nomral_videos}
\end{table}


\begin{table}[]
\caption{Comparison of false alarm rate on abnormal videos of UCF-Crime testing dataset.}
\centering
     \begin{tabular}{|c|c|}
     \hline
\textbf{Algorithms}                     & \textbf{\begin{tabular}[c]{@{}c@{}}False\\ Alarm Rate\\ Abnormal Videos\end{tabular}} \\ \hline
Proposed Method(3D ResNet+constr.+loss) & 0.72                                                                                       \\ 
Proposed Method(3D ResNet+constr.+new rank.loss)    & \textbf{0.67}                                                                              \\ \hline

    \end{tabular}
    
    \label{false_alarm_abnormal_videos}
\end{table}

\section{Conclusions} \label{con}
In this study, we have proposed a deep neural network to detect abnormal activities in the videos. Here, we have used both abnormal and normal videos to train our network. Our proposed model is trained with the help of the deep MIL along with the new proposed ranking loss function. We have validated our proposed algorithm on the UCF-Crime dataset with the help of the ROC-curve and AUC evaluation methods. 
All the experimental results show that our proposed algorithm gives a considerable improvement in decreasing false alarm rates and gives better accuracy in the abnormal activity detection task.
\section*{Acknowledgment}
This work was supported by Institute of Information \& communications Technology Planning \& Evaluation (IITP) grant funded by the Korea government (MSIT) (No.2014-3-00077, AI National Strategy Project).

\bibliographystyle{IEEEtran}
\bibliography{paper}

\end{document}